\documentclass[sigchi]{acmart}
\copyrightyear{2019}
\acmYear{2019} 
\setcopyright{acmlicensed}
\acmConference[ICMI '19]{ICMI '19: ACM International Conference on Multimodal Interaction}{October 14--18, 2019}{Suzhou, Jiangsu, China}
\acmBooktitle{ICMI '19: ACM International Conference on Multimodel Interaction, October 14--18, 2019, Suzhou, Jiangsu, China}
\acmPrice{15.00}
\acmDOI{10.1145/1122445.1122456}
\acmISBN{978-1-4503-9999-9/18/06}
\AtBeginDocument{%
  \providecommand\BibTeX{{%
    \normalfont B\kern-0.5em{\scshape i\kern-0.25em b}\kern-0.8em\TeX}}}

\usepackage{multirow}
\usepackage{multicol}
\usepackage{booktabs}
\usepackage{array}
\usepackage{graphicx}
\usepackage[labelsep=space,
                labelfont={sf,bf},
                textfont=sf]{subfig}
    \usepackage[labelsep=colon,
                labelfont={sf,bf},
                textfont=sf]{caption}

\newcolumntype{P}[1]{>{\centering\arraybackslash}p{#1}}
\newcolumntype{M}[1]{>{\centering\arraybackslash}m{#1}}

\begin{document}

%
\title[Towards Automatic Detection of Misinformation]{Towards Automatic Detection of Misinformation in Online Medical Videos} 

%


\author{Rui Hou}
\email{rayhou@umich.edu}
\affiliation{%
  \institution{Computer Science and Engineering \\ University of Michigan}
  \city{Ann Arbor}
  \state{MI}
  \postcode{}
  \country{USA}
}
\author{Ver\'{o}nica P\'{e}rez-Rosas}
\email{vrncapr@umich.edu}
\affiliation{%
  \institution{Computer Science and Engineering\\ University of Michigan.}
  \city{Ann Arbor}
  \state{MI}
  \country{USA}
}

\author{Stacy Loeb}
\email{stacyloeb@gmail.com}
\affiliation{
    \institution{Department of Urology and Population Health \\ NYU Langone Health}
    \city{New York}
    \state{NY}
    \country{USA}
}

\author{Rada Mihalcea}
\email{mihalcea@umich.edu}
\affiliation{%
  \institution{Computer Science and Engineering \\University of Michigan.}
  \city{Ann Arbor}
  \state{MI}
  \country{USA}
}

%

%
\begin{abstract}

Recent years have witnessed a significant increase in the online sharing of medical information, with videos representing a large fraction of such online sources. Previous studies have however shown that more than half of the health-related videos on platforms such as YouTube contain misleading information and biases. Hence, it is crucial to build computational tools that can help evaluate the quality of these videos so that users can obtain accurate information to help inform their decisions. In this study, we focus on the automatic detection of misinformation in YouTube videos. We select prostate cancer videos as our entry point to tackle this problem. The contribution of this paper is twofold. First, we introduce a new dataset consisting of 250 videos related to prostate cancer manually annotated for misinformation. Second, we explore the use of linguistic, acoustic, and user engagement features for the development of classification models to identify misinformation. Using a series of ablation experiments, we show that we can build automatic models with accuracies of up to 74\%, corresponding to a 76.5\% precision and 73.2\% recall for misinformative instances.
\end{abstract}

%
%


%

\maketitle

\section{Introduction}

The use of social media and other web resources to obtain health-related information is increasing globally \cite{loeb2018prostate,fox2011social}. Video-sharing platforms such as YouTube offer a vast source of medical information on a wide range of medical conditions. Being the largest platform across the world, YouTube has more than 1.9 billion registered users and 5 billion videos watched daily.\footnote{https://www.youtube.com/yt/about/press/} Such large visibility makes YouTube a natural facilitator for the spread of health-related information. Despite its wide coverage of a range of medical conditions and its potential for disseminating health information, health care providers and government agencies alike have expressed concern about the veracity and quality of the information available on this platform~\cite{madathil2015healthcare}.

Health misinformation can be defined as incorrect information that contradicts current established medical understanding, or biased information that covers or promotes only a subset of the facts. Misinformation is harmful not only because it lacks trustworthy sources of medical advice, but  also because it can mislead the audience eventually causing negative effects on their health~\cite{Chou18}. Health misinformation in online medical videos has drawn a large amount of attention recently in the health care community~\cite{drozd2018medical}. Recent research showed that a large number of online medical videos on YouTube are of poor quality and contain misleading or biased contents, with misleading videos even outnumbering high-quality videos on some topics \cite{steinberg2010youtube,madathil2015healthcare,qi2016misinformation,Loeb18Dissemination,Cassidy2018,leong2018youtube}. Therefore, it is essential to ensure the veracity and integrity of health-related online videos and reduce the spread of misinformation so that users can obtain accurate and trustworthy information from the videos~\cite{basch2017content}. However, regular content moderators lack professional medical knowledge to discriminate misinformation from its counterpart, while on the other side it is costly and time-consuming for health professionals to filter out misinformation from the existing massive video repositories. Thus, there is a growing  need  for computational tools to automatically detect misinformation in online medical videos. Surprisingly, very few research studies have addressed this problem.

In this study, we specifically tackle the problem of automatically detecting misinformation in prostate cancer videos on YouTube. According to the American Cancer Society, prostate cancer is the most common cancer among American men and the second leading cause of cancer death, behind lung cancer~\cite{cancerFacts2019}. On YouTube only, there are more than 600,000 videos about prostate cancer, which remain largely unregulated and often contain wrong facts. As a first step to address this problem, we collect a new dataset consisting of 250 videos related to prostate cancer and manually annotate them for misinformation. Using this dataset, we explore the use of linguistic, acoustic, and user engagement features for the development of classification models to identify misinformation. With a series of experiments, we show that we can build automatic models with accuracies of up to 74\%, representing an error rate reduction of 55\% as compared to a majority class baseline.

\section{Related Work}

There have been several efforts to understand the informative value of medical information obtained from search engines and social media platforms versus potential misinformation. Several studies have examined how misinformation disseminates in these platforms on a wide variety of health topics, including vaccination, anorexia, epilepsy, diabetes, and prostate cancer among others~\cite{syed2013misleading,lo2010youtube,leong2018youtube}. Since our study focuses on misinformation on prostate cancer, we mainly discuss previous work in this research direction. However, a more in detail review regarding the impact of misinformation in health care information in social media can be found in~\cite{Lau2012,madathil2015healthcare}.

In 2003, \cite{smith2003internet} conducted a study on evaluating the use of the Web as a source of information by prostate cancer patients undergoing radiotherapy. Authors distributed surveys among 297 patients to evaluate several of internet usages, such as the amount of internet use and the type of internet use. The study found an increased number of patients using the internet to search for information related to prostate cancer. A subsequent study, conducted by \cite{black2006prostate} evaluated the quality of the information available to patients with prostate cancer on the Web. Authors used WebCrawler to query "prostate cancer" and reviewed the top 75 websites by analyzing the degree of coverage and accuracy of the information using 50 criteria aspects. Their findings highlighted the absence of disclosure and attribution of information, which suggests deficiencies in the way information is presented. In \cite{steinberg2010youtube}, 51 YouTube videos on prostate cancer were manually rated by physicians, finding that 73\% of the videos were of fair or poor quality. The work presented by \cite{basch2017content} performed content analysis on the 100 most popular YouTube videos on prostate cancer and emphasized the need for interventions that lead to more accurate online information. Authors in \cite{Loeb18Dissemination} conducted a study on examining misinformation and bias on YouTube evaluating the quality of 150 videos on the screening and treatment of prostate cancer. Their study conducted content evaluation and comparisons between quality, user popularity, and dissemination. The videos were evaluated using instruments such as DISCERN (quality criteria for consumer health information), PEMAT (Patient Education Materials Assessment Tool), and general aspects, including viewer engagement, intended audience, favoring new technology, recommending complementary/alternative medicine and commercial bias. Statistical analyses conducted on these annotations revealed a significant negative correlation between scientific quality and viewer engagement, thus showing that YouTube is a platform that facilitates the dissemination of misinformative content.  All of the above research focused on conducting descriptive and statistical analyses of the content of videos discussing prostate cancer, which confirmed the presence of misinformation in online medical videos and raised important questions regarding users safety and the potential harm associated with the use of poor quality health information. However, different from our work, these studies have not addressed the development of computational tools that support filtering of trustworthy sources of health information in medical videos. 

Furthermore, there have been only a few studies on automatically classifying the quality of health-related content and detecting medical misinformation in social media. Authors in~\cite{liu2018youtube} annotated 600 medical videos on YouTube into two categories: high- and low-medical knowledge videos. A bidirectional LSTM (Long Short-term Memory) model was trained to classify the level of medical knowledge encoded in the collected videos using word-vector representations extracted from the video descriptions. Their study also presented comparisons between text-based methods and lexicon-based medical extraction and showed that the former is a promising method on classifying the level of medical knowledge contained in YouTube videos. 

The work in~\cite{ghenai2018fake} aimed to automatically identify Twitter users who post health misinformation by comparing 4,212 users who promote ineffective cancer treatments with the same number of general users interested in cancer. Logistic regression with LASSO (Least Absolute Shrinkage and Selection Operator) regularization was trained based on linguistic features extracted from the LIWC (Linguistic Inquiry and Word Count) lexicon and text readability and found that readability, tentative language and avoidance of personal pronouns are important language markers associated to the propensity of posting cancer treatment misinformation. 

Overall, these studies have examined the quality of health information presented in social media as well as the characteristics of individuals who spread misinformation in such sources. However, we are not aware of previous research on the automatic detection of misinformation in health-related videos, as we propose in this work.

Finally, also related to our research are the studies on automatic deception detection, particularly those focused on the identification of deceptive content in social media such as fake news and rumor detection~\cite{perez2017automatic,shu2017fake,boididou2017learning,krishnamurthy2018deep}. Research in these areas has shown that using multi-modal features, including social media user engagement, linguistic, visual and acoustic features can help improve the accuracy on automatically detecting deceptive or inaccurate content. Our work draws inspiration from this previous work to derive several learning features that attempt to capture misinformative behavior in YouTube videos. 

\section{Misinformation Dataset}
We leverage health-related videos for the development of a misinformative video detection model. More specifically, we focus on videos available on YouTube that are related to prostate cancer and construct a dataset consisting of 250 videos manually annotated for misinformation.

\subsection{Data Collection}

Our data collection consisted of two steps. First, we attempted to identify trustworthy content on prostate cancer by conducting searches using keywords such as prostate cancer screening and prostate cancer treatment. Second, we conducted similar searches but this time adding keywords that could potentially trigger misinformative content, such as prostate cancer miracle cure, and prostate cancer natural remedies. 

Our search yielded a large set of videos discussing various aspects of prostate cancer, including risk factors, prevention, screening, treatment options, and side effects. The collected videos are contributed both by individual users (e.g., patient narratives, healthcare professionals), reputable healthcare organizations (e.g., academic medical centers, professional medical societies), media groups and commercial entities. 

Second, we manually filtered the obtained videos using the following five criteria: (1) the primary content should be about prostate cancer; (2) the video should not contain animations or artificial voices; (3) the language of the video should be English; (4) the duration of the video should be less than 30 minutes; (5) the video should not contain more than two speakers.\footnote{About 40\% of the videos in our dataset have two speakers. These videos mainly consist of interviews with doctors and patients.} After this screening, a total of 250 distinct videos were selected and downloaded. 
These constraints are motivated by the need for videos with a reasonable audio-visual quality so that multimodal feature extraction can be applied successfully. The collected 250 videos have collectively gained more than 10 million views as of April 2019, thus reaching a large audience, which again emphasizes the importance of ensuring the integrity of health-related content on YouTube. 

\noindent 
\textbf{Privacy Considerations.} To address ethical and privacy concerns during the data collection, we only accessed videos listed under the Creative Commons Attribution license (reuse allowed) before conducting the analyses described in this paper.

\subsection{Data Annotation}

We evaluated the extent of misinformation contained in the videos using a 5-point Likert scale, with 1 indicating that the video contains no misinformation and 2-5 indicating an increasing level of misinformation. This is in line with the scoring system employed by DISCERN, an educational tool designed primarily to enable patients (or health consumers), their carers and advisers to select and use written information on treatment choices as part of good quality healthcare~\cite{charnock1999discern}. This tool also has been previously applied to the evaluation of YouTube videos~\cite{cassidy2018youtube}. Thus, we deemed a video misinformative if it contains guideline-discordant information (e.g., recommends a treatment that is not recommended in the guidelines), heavily relies on anecdotal evidence (e.g., patient narrative asserting that use of dietary ingredient cured his prostate cancer without scientific evidence), or if it contains advertisements or biased content (e.g., promoting a certain therapy without describing risks or alternatives).

The videos that presented the guideline-concordant and correct medical information in an objective, unbiased, and scientific manner were considered trustworthy. For example, a trustworthy video regarding prostate cancer treatment tends to introduce multiple available treatment options instead of heavily promoting just one treatment, or discusses side effects and risks alongside the benefits. 

\begin{table}[t]
  \caption{Distribution of Misinformation Scores}
  \label{tab:dist_level}
  \begin{tabular}{ccc}
    \toprule
    Misinformation Score & \# Videos & Percentage \\
    \midrule
     1  & 132 & 52.8\% \\
     2 & 43 & 17.2\% \\
     3  & 38 & 15.2\% \\
     4 & 26 & 10.4\% \\  
     5  & 11 & 4.4\%\\
  \bottomrule
\end{tabular}
\end{table}

The coding was conducted by several expert urologists who watched each video and assigned the corresponding score. The within-one inter-annotator agreement on a 10\% random sample of the videos\footnote{Given the relatively small data sample with dual annotations, instead of measuring correlation, we measured a ``relaxed'' agreement, where annotations within a range of one were considered in agreement.} was measured at 92\%, thus indicating a high level of agreement among annotators. 

Table~\ref{tab:dist_level} shows the distribution of the misinformation scores assigned to the videos in the dataset. Because of the skewed frequency distribution, we decided to conduct our analysis using a binary categorization. Thus, videos with a misinformation score of 1 are categorized as trustworthy and those with misinformation scores ranging from 2 to 5 are categorized as misinformative. This resulted in a video dataset consisting of 132 trustworthy videos and 118 misinformative videos. The average duration of each type of video is 297 and 324 seconds respectively, with misinformative videos being slightly lengthier than their counterparts.

\subsection{Transcriptions}
We obtained video transcriptions using the YouTube automatic captioning service. For 33 videos for which auto-generated captions were not readily available, we used the Speech-to-Text service provided by Google Cloud to transcribe their corresponding audio. Since automatic speech recognition produces unpunctuated text, we applied automatic punctuation restoration to improve the transcription readability and also to enable further language analysis that might depend on punctuation. We used the bidirectional recurrent neural network with attention mechanism introduced in ~\cite{tilk2016}, pre-trained on 40 million words from the Europarl v7 English corpus \cite{koehn2005europarl}. 
Note that the algorithm used for automatic punctuation restoration does introduce some degree of error, but this error is small enough so as not to affect the subsequent text processing steps. 

The final transcription set consists of 193,059 words, with an average of 772 words per transcript. Table ~\ref{tab:sample_transcript} shows sample transcript excerpts for misinformative and trustworthy videos.

\begin{table*}[t]
    \caption{Sample transcript excerpts for misinformative and trustworthy prostate cancer videos}
    \label{tab:sample_transcript}
    \centering
    \begin{tabular}{m{8cm} | m{8cm}}
    \toprule
    \multicolumn{1}{M{8cm} | }{Misinformative} & \multicolumn{1}{M{8cm}}{Trustworthy} \\
    \midrule
       How I healed this prostate cancer, naturally. I was diagnosed back in 2012. My urologist wanted to have me cut it, cut the prostate out. I was just freaked me out. I chose the alternative route. There's lots of research with a guy named Don Tolman? Do a Google search on him? He is amazing: lots of raw fruit and vegetables. Lots of water, fasting sunshine, fresh air, exercise, lots of turmeric and curcumin. No doubt about that, and the miracle cure, as I mentioned before, you can get into the description. Click on the link to the miracle cure video and have a look. It really is amazing. It is a miracle for sure the benefits of the miracle cure on my body along there, cancer free, no medical intervention at all. I'm healthier than I've ever been.  &  Some prostate cancers are low-grade non-aggressive and they may grow very very slowly for many many years and they may require no treatment in that situation. We will monitor the PSA over time, and that is called active surveillance. In patients who have a higher grade of tumor, we may offer them either some form of radiotherapy treatment or some form of surgery. The radiotherapy treatments lie between brachytherapy, where we put radioactive seeds individually into the prostate, to kill a prostate cancer in situ or external beam radiotherapy where the radiation is beamed in from outside the body. In both of these treatments, the prostate gland remains in place in the body. Surgery requires removal of the prostate gland. \\
    \bottomrule
    \end{tabular}
\end{table*}

\subsection{Preprocessing}
We conducted a set of preprocessing steps to enable the automatic text and audio feature extraction. First, we processed the transcripts by applying tokenization, part-of-speech (POS) tagging, lemmatization, and dependency parsing. We used the Stanford CoreNLP package as our primary tool set for these tasks \cite{manning-EtAl:2014:P14-5,toutanova2003feature,chen2014fast}. Second, we separated the audio from the video and converted the signal to a single channel and to a uniform sample rate of 16k using ffmpeg.\footnote{https://ffmpeg.org/}

\section{Automatic Detection of Misinformation}

We conduct several learning experiments that aim to classify the videos as misinformative or trustworthy. We extract several feature sets to capture the language and acoustic cues of misinformative content. Our features incorporate aspects that were previously found related to misinformative content such as viewer engagement, i.e., number of views, thumbs up, thumbs down,  number of comments, and statistics of the video duration~\cite{Loeb18Dissemination,qi2016misinformation}. We also extract standard language features such as ngrams and syntax-based features to capture the differences between misinformative and trustworthy language. In addition, since misinformation is a problem related to fake news content, we also experiment with linguistic features that have been found useful in the identification of this type of content, including readability, lexical richness, and psycholinguistic features derived from the LIWC lexicon~\cite{Jiang:2018:LSU:3290265.3274351}. Furthermore, we also explore whether the speech signals can provide additional information to discriminate between the types of content by extracting a large set of raw acoustic features.

During our experiments, we first explore the predictive power of each feature set separately and also build a model that integrates all features. 

The experiments are conducted using a linear kernel Support Vector Machine (SVM) classifier as the main classifier. We use the implementation available in scikit-learn \cite{scikit-learn}. More specifically, the implementation we choose is the LinearSVC class based on the LibLinear library \cite{Fan:2008:LLL:1390681.1442794}. We normalize each feature using L2 norm before feeding it into the classifier. 

We selected an SVM classifier due to the limited number of examples and the large number of features. SVM classifiers have shown good performance in multimodal tasks in the past. We experimented with other algorithms, however, they did not perform better than the SVM classifier. Additionally, due to the limited training data we were unable to apply successfully a deep learning-based model.

Our evaluations are conducted by doing five-fold cross-validation with accuracy, precision, recall, and F1-score as performance metrics.  We use a majority class baseline as the reference value for our experiments. This value is obtained by selecting \textit{trustworthy} as the default class, which corresponds to a 52.8\% accuracy.   

\subsection{Viewer Engagement Features} This feature set contains four viewer engagement attributes available for most of the videos on YouTube as well as two descriptive metrics. These include: 1) the average number of video views per day; 2) the number of comments posted below the video; 3) the number of thumbs up; 4) the number of thumbs down; 5) the video duration in seconds; and 6) the video category. We calculate the average number of views per day starting from the publication date up to the date of the query.  The video category is usually assigned by the video owners themselves, among 32 available categories. A few examples of the categories in our dataset are people and blogs, education, science and technology, and news. 

\begin{figure*}[t]\centering{
    \subfloat[Thumbs up ]
        {\includegraphics[height=0.15\textheight, width=0.32\textwidth]{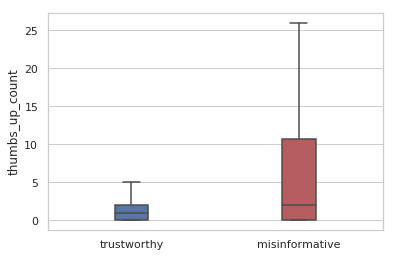}}
    \hfill
    \subfloat[Thumbs down]
         {\includegraphics[height=0.15\textheight, width=0.32\textwidth]{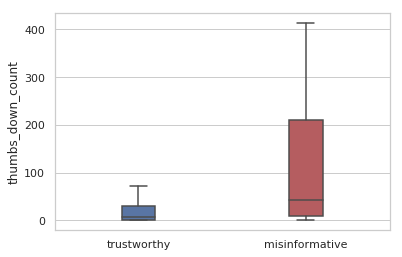}}
    \hfill
    \subfloat[Average views per day ]
         {\includegraphics[height=0.15\textheight,width=0.32\textwidth]{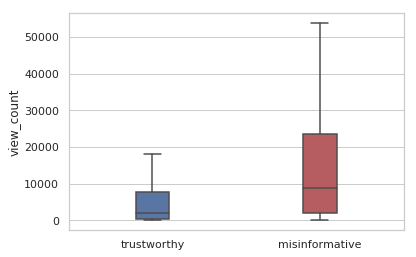}}}
        \hfill
    \subfloat[Number of comments ]
        {\includegraphics[height=0.15\textheight,width=0.32\textwidth]{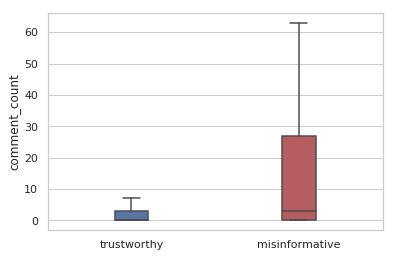}}
    \subfloat[Video duration]
        {\includegraphics[height=0.15\textheight,width=0.32\textwidth]{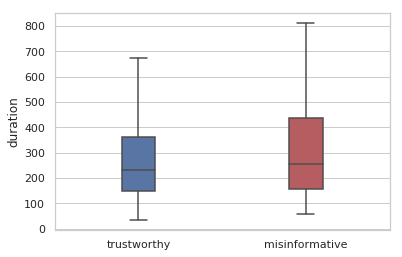}}
    
    \caption { YouTube viewer engagement features for misinformative and trustworthy videos}
    \label{fig:viewer_eng}
    \end{figure*}

All six attributes are obtained using the official YouTube API.\footnote{\url{https://developers.google.com/youtube/v3/}. The different attributes are retrieved as of April, 18, 2019.} Note that we encountered 14 videos with comments disabled, and in that case, the corresponding features were set to zero. Figure~\ref{fig:viewer_eng} shows the distribution of the different features for both misinformative and trustworthy videos. The plot suggests significant differences between the two groups for these attributes, thus potentially making them good discriminators for the two types of content. 

\subsection{Linguistic Features}
We extract several sets of linguistic features derived from the video transcript to model misinformative language. The features are extracted as follows:

\noindent
\textbf{Ngrams.}
We extract unigrams and bigrams i.e., all the unique words and word-pairs present in the transcript, using the bag of words representation of the tokenized transcriptions. Considering a large number of distinct ngrams that exist in all of the transcripts, we set a frequency threshold of 10, which was obtained experimentally on a development set. We represent these features using tf-idf (term frequency-inverse document frequency) values to account for the importance of each ngram and the variance in transcript length.

\noindent
\textbf{Readability.}
 This feature set aims to measure how easy it is to understand the content of the video. We use several metrics for sentence and token usage, including: (1) Fourteen features on sentence's usage, including the number of sentences, the number of words per sentence, the number of syllables, the number of word types, the number of long words; (2) Six features describing the overall word usage such as the number of pronouns and prepositions; and (3) Six features that quantify sentence beginnings, i.e., the number of subordinations and conjunctions. We also consider nine standard readability metrics, including the Automatic Readability Index (ARI), Flesch-Kincaid Grade Level, Flesch Reading Ease, Coleman-Liau, Gunning Fog Index, LIX, Simple Measure of Gobbledygook (SMOG) Index, RIX and Dalechall Index. The above 35 features are extracted using an open-source readability toolkit.\footnote{\url{https://pypi.org/project/readability/}}

\noindent
\textbf{LIWC.} 
This set of features consists of 73 word classes present in the 2015 version of the LIWC lexicon, which includes psycho-linguistic processes (e.g., affect words, social words), psychological processes (e.g., cognitive process, perceptual processes, core needs, and drives) as well as grammar and punctuation variables. The features are encoded as the percentage of each LIWC class present in the transcript.

\noindent
\textbf{Syntax.} 
These features consist of language production patterns based on the probabilistic Context Free Grammar (CFG) parse trees of the transcription \cite{feng2012syntactic}. The features are derived from the transcription parse tree, taking into account all the lexicalized production rules (including rules with terminal nodes) and combine them with their grandparent node. An example of such a feature can be *NN$\land$NP$\rightarrow$diagnosis where NN (i.e., noun) is the grandparent node, NP (i.e., noun phrase) is the parent node, and \textit{diagnosis} is the terminal node. Like ngrams, we also encode them as tf-idf values with a frequency threshold of 10. 

\noindent
\textbf{Lexical Richness. }

This feature set measures the lexical richness~\cite{lu2012relationship} of the transcription. It includes metrics such as lexical density, i.e, the ratio of lexical words to the total number of words in the transcript; lexical sophistication, i.e., the proportion of unusual or advanced words in the transcript; and lexical variation, represented as the ratio of word types to the number of words in the transcript.

\begin{table*}[t]
  \centering  
  \small
  \caption{Misinformation classification results using  viewer engagement features, linguistic and acoustic features and their combination.}
  \label{tab:result_1}
  \begin{tabular}{llcccccccc}
  \toprule
  \multirow{2}{*}{Category} & \multirow{2}{*}{Feature set} & \multirow{2}{*}{\# Features} & \multirow{2}{*}{Accuracy} & \multicolumn{3}{c}{Misinformative} & \multicolumn{3}{c}{Trustworthy} \\
    & & & & Precision & Recall & F1-score & Precision & Recall & F1-score\\ 
    \midrule
    \multicolumn{2}{l}{Majority baseline} &  &  52.80\% & 0\% & 0\% & 0\% & 100\% & 100\% & 100\% \\
    \midrule
    Youtube                    
    & (1) Viewer engagement & 6 & 61.56\% & \textbf{96.00\%} & 21.05\% & 31.09\% & 58.56\% & \textbf{97.78\%} & 73.01\% \\
    \midrule
    \multirow{6}{*}{Linguistic}               
    & (2) LIWC & 73 & 67.62\% & 68.89\% & 61.30\% & 62.77\% & 69.79\% & 73.30\% & 70.03\% \\ 
    & (3) Ngrams & 3577 & 71.61\% & 74.00\% & 68.95\% & 68.11\% & 75.63\% & 74.02\% & 72.38\% \\
  
    & (4) Lexical richness & 33 & 48.78\% & 27.00\% & 13.33\% & 16.89\% & 50.94\% & 80.48\% & 61.99\% \\
    & (5) Syntax (CFG) & 3270 & 70.41\% & 73.54\% & 67.28\% & 67.15\% & 74.27\% & 73.30\% & 71.22\% \\
    & (6) Readability & 35 & 57.63\% & 57.66\% & 40.94\% & 46.40\% & 58.23\% & 72.62\% & 64.01\% \\
    \cmidrule{2-10}
    & All linguistic & 6988 & 72.41\% & 75.27\% & 72.28\% & 70.96\% & 76.29\% & 74.07\% & 72.93\% \\ 
    \midrule
    \multirow{3}{*}{Acoustic}
 & (8) Emo\_IS09 & 384 & 58.48\% & 57.85\% & 47.71\% & 51.06\% & 60.02\% & 68.23\% & 63.28\% \\
    & (7) Emobase & 989 & 53.63\% & 52.78\% & 46.81\% & 48.19\% & 55.69\% & 59.89\% & 56.95\%  \\
    & (9) Emo\_large & 6552 & 57.17\% & 55.31\% & 51.07\% & 52.10\% & 59.32\% & 62.74\% & 60.21\% \\ \hline
\multirow{2}{*}{Combined} 
 & (1)+(3)+(8) & 3968 & 72.39\% & 76.36\% & 68.91\% & 68.96\% & 75.59\% & 75.53\% & 73.32\% \\ 
 & (1)+Ling+(8) & 7379 & \textbf{74.41\%} & 76.51\% & \textbf{73.15\%} & \textbf{71.93\%} & \textbf{78.44\%} & 75.58\% & \textbf{74.86\%} \\ \bottomrule
\end{tabular}
\end{table*}

\subsection{Raw Acoustic Features} 
The last category of features consists of a large number of raw speech features extracted with the  openEAR~\cite{eyben2009openear} toolkit. OpenEar can extract low-level audio features such as loudness, pitch, and twelve mel-frequency cepstral coefficients (MFCC) and also perform various types of statistical manipulation to generate more advanced features. We experiment with three predefined sets of features that are frequently used for emotion recognition tasks: emobase containing a baseline set; emo\_IS09, which was used for the Interspeech 2009 Emotion Challenge; and emo\_large, which contains an extended set of low-level audio features and post-processing functionals from the emobase baseline set.

\subsection{Classification Results}

Classification results obtained with each feature set are shown in Table~\ref{tab:result_1}, along with the number of features per set. From this table, we observe that the different linguistic features achieve accuracies well above the baseline, except for the lexical diversity features, which show slightly lower accuracies. Among all linguistic features, the ngrams perform the best, followed by the syntax features and the LIWC features. Furthermore, the model that integrates all linguistic features outperforms each individual set, with improved precision, recall and F-score metrics. 

For the acoustic features, the Emo\_IS09 set obtains the best performance among the three acoustic feature sets. Interestingly, this set contains a smaller number of features thus suggesting that features in the other sets might be redundant. Due to its performance, we selected Emo\_IS09 as the main feature set representing the raw acoustic characteristics in the videos for further experiments. Note that we did not test the three acoustic feature sets jointly as they are incremental. 

We also evaluated two integrated classifiers that use features from different modalities. The first model is built using the viewer engagement features, all linguistic features, and the Emo\_IS09 acoustic set, while the second model includes the viewer engagement features, ngrams only, and Emo\_IS09. The first model, which combines all features, reaches the highest accuracy of 0.74 and the highest F1-scores corresponding to the two classes, 0.72 and 0.75, respectively. Furthermore, the integrated model outperforms models built using a single modality, thus suggesting that the different feature sets from the different modalities contribute complementary information.

\section{Discussion}

 Previous studies examining the quality of information presented in online videos on prostate cancer were conducted by applying statistical analyses on a small set of manually annotated features. To the best of our knowledge, our work is one of the first to address the automatic classification of misinformative content in medical videos posted in online sources. 
 
 We draw inspiration from previous work in the medical field to build a dataset of medical videos from YouTube, annotated with misinformation \cite{Loeb18Dissemination}. We believe that the collected dataset can be a valuable resource to enable further research on the health misinformation phenomena, particularly for detecting misinformation in online videos, a research venue that is currently underexplored. While the constraints we impose during the data collection might limit the diversity of the dataset collected, they also have the purpose of generating a dataset that reflects the multimodal nature of the dataset, without adding additional complexity (i.e., dealing with multiparty conversations). Hence, they enable multimodal analyses that focus on exploring the properties of misinformation rather than multimodal aspects of the collected videos. We believe that our method will generalize well to the large number of videos that have the same properties as the videos in our dataset. 
 Our future research will consider expanding our data collection to analyze a larger sample and confirm our findings
 
 The presented analyses are focused on the verbal and non-verbal channels as well as the user engagement; however, the dataset can be further analyzed to extract additional features from modalities such as the visual channel. Furthermore, our study addresses misinformation detection as a binary classification task but since the data contains fine-grained annotations for the level of misinformation present in the videos, future research can explore this task on several levels to build ad hoc models. 
 
 Our preliminary analyses in this new dataset show promising results by successfully distinguishing between misinformative and trustworthy content with accuracies well above the majority class baseline. We explore several sets of linguistic, user engagement and acoustic features individually and jointly. The experiments with individual feature sets show that they can capture cues that differentiate between misinformative and trustworthy content. In particular, the joint model showed that integrating features from multiple modalities is beneficial for this task. This work uses early fusion as the main strategy to combine the different features, future work could explore more advanced methods to better capture the correspondence between modalities. 
 
 Despite the promising results, our work has also several limitations. First, the restrictions we impose during the data collection made us overlook an important fraction of misinformative videos that consisted mainly of automatic narratives with animations and voice-overs. These videos could potentially contain other cues to misinformation such as the presence of animations or the use of graphical aids. In order to make use of this data, future research will require computational methods that are able to deal with incomplete and noisy modalities. Similarly, the data collection restrictions favored the inclusion of videos portraying laypeople i.e., patients who share anecdotal experiences with prostate cancer, thus introducing some degree of bias in terms of lay users sharing mostly unscientific information. Second, only a small sample of the videos was doubly annotated by medical experts due to the laborious nature of this task; therefore a future direction is a need for a larger number of videos to be reviewed by multiple medical experts to verify the extent of misinformation that is present by consensus. Third, the linguistic feature extraction is conducted on transcripts that were obtained using automatic transcription and punctuation restoration, thus introducing noise during this process, which in turn could affect the performance of the classifiers. We consider exploring low dimensional representations that do not rely on punctuation or grammatical correctness such as word embeddings as a future research venue.

\section{Conclusions}

In this paper, we explored the task of automatically identifying misinformation in online medical videos. We constructed a dataset consisting of 250 YouTube videos related to prostate cancer, manually categorized as being trustworthy or misinformative. We then experimented with linguistic, acoustic, and user engagement modalities, and extracted several sets of features to train SVM classifiers for misinformative video detection. Our results showed that following a multimodal approach by integrating features from different modalities outperformed individual modalities reaching an overall accuracy of 74\%. 

\begin{acks}
This material is based in part upon work supported by the Michigan Institute for Data Science, by the National Science Foundation (grant \#1815291), by the John Templeton Foundation (grant \#61156), by DARPA (grant \#HR001117S0026-AIDA-FP-045), by the Prostate Cancer Foundation, and by the Edward Blank and Sharon Cosloy-Blank Family Foundation. Any opinions, findings, and conclusions or recommendations expressed in this material are those of the author and do not necessarily reflect the views of the Michigan Institute for Data Science, the National Science Foundation, the John Templeton Foundation, DARPA, the Prostate Cancer Fundation , or the Edward Blank and Sharon Cosloy-Blank Family Foundation. 
\end{acks}

\bibliographystyle{ACM-Reference-Format}
\bibliography{refs}


\end{document}